\documentclass[11pt,a4paper]{article}
\usepackage[hyperref]{emnlp2018}

\usepackage{latexsym}
\usepackage{hyperref}  % NOTE(lucypark): for \autoref
\usepackage{subcaption}  % NOTE(lucypark): for subfigures
\usepackage{lipsum}% http://ctan.org/pkg/lipsum

\usepackage{graphicx}% http://ctan.org/pkg/graphicx
\usepackage{multirow, makecell}
\usepackage{amsmath}
\usepackage{amsfonts}  % NOTE(lucypark): for mathbb
\usepackage{float}
\usepackage{tabularx}
\usepackage[title]{appendix}

\usepackage{url}

\title{Text Classification using Capsules}

\author{Jaeyoung Kim, Sion Jang \and Sungchul Choi \\ TEAMLAB, Gachon University \\  teamlab.gachon@gmail.com
        \AND
         Eunjeong Park \\ NAVER \\ lucy.park@navercorp.com
}

\date{}

\begin{document}
\maketitle
\begin{abstract}
This paper presents an empirical exploration of the use of capsule networks for text classification.
% We applied the model used in image classification as a model of text classification.
% We have shown through experimentation that capsule networks have potential for text classification.
While it has been shown that capsule networks are effective for image classification, their validity in the domain of text has not been explored.
In this paper, we show that capsule networks indeed have potential for text classification,
and that they have several advantages over convolutional neural networks.
% Compared to convolutional neural networks, we have identified several advantages of capsule networks.
We further suggest a simple routing method that effectively reduces the computational complexity of dynamic routing.
% Our model structure is very simple and effectively reduces the computational complexity of routing, which was one of the main disadvantages of in the original capsule networks.
% We utilized seven benchmark datasets for text classification and confirmed the benefits of capsule network.
We utilized seven benchmark datasets to demonstrate that capsule networks, along with the proposed routing method provide comparable results.
\end{abstract}

\section{Introduction}
Text classification is one of the most basic and important tasks in the field of machine learning.
% One of the most important aspects in text classification is the representation of words and documents.
Traditionally, the use of
term frequency inverse document frequency (tf-idf) as a representation of documents,
and general classifiers such as support vector machines (SVM)
% NOTE: fan2008liblinear does not compile
%SVM~\cite{joachims1998text, fan2008liblinear}
or logistic regression have been utilized for statistical classification.
% ~\cite{joachims1998text}.

% Recently, however, continuous development of deep learning methods has made it possible to efficiently embed words into a continuous space~\cite{mikolov2013distributed}.
Recently, however, continuous development of deep learning methods has made it possible to find distributed representations of words and documents in an efficient manner~\cite{mikolov2013distributed, le2014distributed},
which further led to higher accuracies for text classification.
The major deep learning models utilized in text classification are largely based on convolutional neural networks (CNNs) and recurrent neural networks (RNNs).
% , and long short-term memory (LSTM)~\cite{hochreiter1997long}.
% and text classification CNN(Kim, 2004) proved that CNN works efficiently in text classification. And Character-level convolutional networks (Zhang Xiang., 2015) performs well in large data sets.
% Recently, however, a capsule network(Sara Sabour et al., 2017) has announced that has complemented the drawbacks of CNN in image classification and presented a new direction. and capsule networks achieves state-of-the-art performance on MNIST(LeCun et al., 1998) i of benchmark data sets of image classification.

% Meanwhile, in the image classification domain, capsule networks~\cite{hinton2011transforming, sabour2017dynamic} have overcome several drawbacks of CNNs and proved to be effective at understanding spatial relationships in high levels of data.
Meanwhile, in the image classification domain, capsule networks~\cite{hinton2011transforming, sabour2017dynamic}
proved to be effective at understanding spatial relationships in high levels of data
by employing a whole vector of instantiation parameters.
We have applied this network structure to the classification of text,
and argue that it also has advantages in this field.

The main contributions of this work are three-fold.
First, we apply capsule networks with dynamic routing to text classification and achieve comparable results to previous methods.
Second, we propose an alternative routing method that achieves higher accuracy compared to dynamic routing.
Third, we propose the use of an ELU-gate~\cite{dauphin2016language} to propagate relevant information.

\section{Related Work}

\subsection{Text classification}
%\cite{zhang2015character}
%\cite{kiros2015skip}
%\cite{silva2011symbolic}
%\cite{kokkinos2017structural}

As deep learning architectures have become more popular, they have also been applied to text classification.
CNN models were originally popularized for text classification by~\newcite{kim2014convolutional} and employed convolutions directly to sentences.
CNNs were further explored at the character-level by~\newcite{zhang2015character}.
Dynamic convolutional neural networks (DCNNs)~\cite{kalchbrenner2014convolutional} introduce a unique method of pooling by dynamically incorporating the length of a sentence when determining the pooling parameter.

While it is straightforward to utilize RNNs for text classification because of the sequential nature of text,
naive RNNs have not been as successful as anticipated.
However, with long short-term memory (LSTM) and initializations based on sequence autoencoders~\cite{dai2015semi}
or small perturbations added to LSTM word embeddings~\cite{miyato2016adversarial},
RNNs have also achieved strong results.

Additionally, self-attention networks
- models without any convolutions or recurrence -
have also been successfully applied to text classification~\cite{shen2018bi}.

\begin{figure*}[htbp!]
  \includegraphics[width=\textwidth]{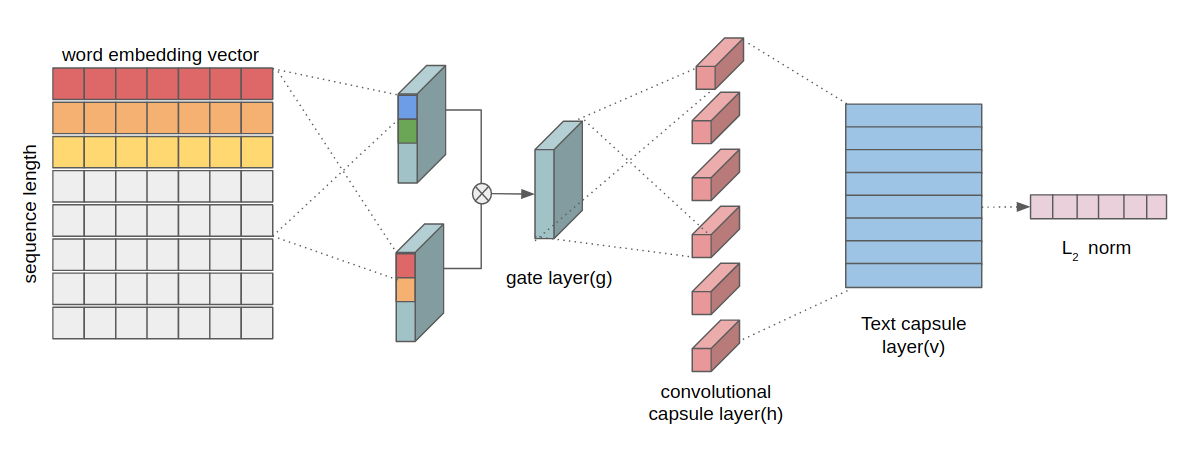}
  \caption{Capsule networks for text. Each document passes a gate layer, convolutional capsule layer, and a text capsule layer.}\label{fig:model_architecture}
\end{figure*}

\subsection{Capsule networks}

Because the convolution operator in a CNN is represented by a weighted sum of lower layers, it is difficult to express the features of a complex object as it moves into the upper layers. This has the disadvantage of not considering the hierarchical relationships between local features.
CNNs utilize pooling to overcome these shortcomings. Pooling can reduce the computational complexity of convolution operations and capture the invariance of local features.
However, pooling operations lose information regarding spatial relationships and are likely to misclassify objects based on their orientation or proportion.

The capsule network is a structured model that solves many of the problems inherent to CNNs.
Capsules in capsule networks are locally invariant groups that learn to recognize the existence of visual entities and encode their properties into vectors.
While neurons operate independently in a CNN, capsule networks utilize a nonlinear function called squashing because capsules (groups of neurons) are represented as a vector.
% FIXME: the below sentence
% This squashing operation represents the probability that an object exists while maintaining a vector orientation.
% Capsule networks~\cite{sabour2017dynamic} attempt to conserve positional information of features as the complexity of CNN layer increases.

Capsules consider the spatial relationships between entities and learn these relationships via dynamic routing~\cite{sabour2017dynamic}.
%And upper entity represented by reasonal combination of lower-level capsules using Dynamic routing.
% In dynamic routing, the link strength is updated according to the similarity between the vector predicting the lower capsule and the upper capsule.
% CNN used invariance to extract features of the image using max-pooling, while Capsule networks tried to solve the classification problem using equivariance.
%It is told that Capsule networks pursue equivalence through dynamic routing, and
Dynamic routing determines the connection strength between lower-level and upper-level capsules through repetitive routing based on a coupling coefficient.
This coupling coefficient is utilized to measure the similarity between the vectors that predict the upper capsule and lower capsule, and learns which lower-level capsule must be directed to which upper-level capsule.
Through this process, capsules learn to represent the properties of a given entity.

% \subsection{Gated Convolution Networks}

\section{Model}
\subsection{Architecture}

%Our aim is apply Capsule networks in text classification, changed according to purpose.
Our goal is to apply capsule networks to text classification, and modify it according to our purpose.
% This method, called capsule, can represent attributes of partial entities. These capsules also use the routing technique to make the sub-capsules reasonably select the upper capsules. Also, it has the advantage of expressing the meaning in a wide space by expressing the entities expressed by the existing scalar as a vector.
Capsules have the ability to represent attributes of partial entities, and express semantic meanings in a wider space by expressing the entities with a vector rather than a scalar.
In this regard, capsules are suitable to express a sentence or document as a vector.
%Our model is very simple and shows the possibility of capsule.
%Figure \ref{fig:model_architecture} is the structure of the model.
%The input of the network is a document D, which is a sequence of
\autoref{fig:model_architecture} depicts the general structure of the proposed model.
The input of the network is a document \(D\in \mathbb{R}^{l\times e}\),
% which is represented as
%
% \begin{equation}
% D = \{ w_l | \, l \in ( 1, l ) \}, w_l\in R^e,
% \end{equation}
where \(l\) is the length of the document and \(e\) is the embedding size.
%We designed a model that starts with a trained word vector. Trained word vectors used 100 billion words of Google News data\footnote{\url{https://code.google.com/archive/p/word2vec/}}.

% Our model uses pretrained word vectors, consisting of 100 billion words from Google News.\footnote{\url{https://code.google.com/archive/p/word2vec/}}.
% We have experimentally found that using a fixed filter size is more efficient than connecting multiple filter sizes, such as \(f\in(2,3,4,5)\) in our model.
% We used a filter size of \(f\) in datasets.

The second layer is a feature map utilizing convolutions, where the kernel size is \(f\times e\), number of filters is \(n\),
and stride is fixed to 1.
While CNNs utilize max-pooling of feature maps to extract meaningful contexts,
% This can reduce the number of learning parameters to create an efficient model and extract context semantically.
% FIXME: 2.2에 elu 없을 때의 공식도 추가해야할듯.
we utilize a trick similar to the gated-linear unit~\cite{dauphin2016language}, defined as

\begin{equation}
    g(D) = (D * W + b)\otimes elu(D * V + c),
\end{equation}

where \(W, V \in \mathbb{R}^{f\times e\times n}\) are weights, \(b, c\) are bias terms, and \(\otimes\) is the element-wise multiplication operator.
This ELU-gate unit acts as a control tower by selecting which features to be activated.
% and activating meaningful contexts in the feature map.
% Similar to LSTMs, these gates control the information that is passed through the hierarchy.
% TODO: 그런가? 왜인지 설명 필요
Unlike pooling, the ELU-gate unit does not lose spatial information.
% And this help the vanishing gradient problem.

The next layer \(h\)
is a convolutional capsule layer
with $a$ channels of convolutional \(M\) dimension capsules
where the kernel size is \((l-f+1)\times 1\).
%왜 large kernel size를 사용했는지
% 1. capsule의 개수를 줄이기 위해서
% 2. valid receptive field를 넓히기 위해서
% FIXME: clarify the below sentence
Because the classifier is connected locally to the feature map, it is difficult for the classifier to handle variations in transformation.
Some studies have shown that utilizing a large kernel size in a network tends to gather information from a much larger region in the receptive field~\cite{peng2017large}.
Because we do not utilize pooling, we instead increased the size of the kernel to enlargen our viewpoint.
% FIXME: therefore?
Therefore, we applied nonlinear squashing~\cite{sabour2017dynamic} in the convolutional capsule layer \(h\).

The final layer is the text capsule layer $v$.
We utilized two different routing methods from the convolutional capsule layer to the text capsule layer, as described in the subsections below.
% One is CapsNet-routing, where dynamic routing is employed,
% and the other is CapsNet-non-routing, which is capsules without routing.

\subsubsection{Capsule network with dynamic routing}
In \newcite{sabour2017dynamic},
the capsule network updated the weight of coupling coefficients through an iterative routing process and determined the degree to which lower capsules were directed to upper capsules.
%capsules in the convolutional capsule layer \(h\) select the upper capsule through dynamic routing.
The coupling coefficient is determined by the degree of similarity between the standard-upper and prediction-upper capsules.

\begin{equation}
 c_{ij} = \frac{\exp (b_{ij})}{\sum_k \exp (b_{ik})}
\end{equation}

where \(i\in [1,a] , j\in [1, k]\), and \(k\) is the number of classes. \(c_{ij}\) is the coupling coefficient and the softmax output of \(b_{ij}\) is updated in every routing iteration. \(b_{ij}\) is determined by the degree of similarity between the lower and upper capsules and predicts the entities of the upper capsules. The predicted vector \(\hat{h}_{j|i}\) is expressed by a matrix operation between the weight matrix \(W_{ij}\) and \(h_i\).

\begin{equation}
  \begin{aligned}
    \hat{h}_{j|i} &= W_{ij} h_i \\
    W_{ij} &= [M\times N]
  \end{aligned}
\end{equation}

The routing procedure is defined as follows:

\begin{equation}
\label{eq:routing}
  \begin{aligned}
 s_j = \sum_{i} c_{ij}\hat{h}_{j|i} \\
 v_j = \frac{||s_j||^2}{1+||s_j||^2}\frac{s_j}{||s_j||} \\
 b_{ij} \gets b_{ij} + v_j\cdot\hat{h}_{j|i}
 \end{aligned}
\end{equation}

\subsubsection{Capsule network with static routing}
% NOTE: Our conjecture is that equivariant한 manifold가 language에서는 좀 더 flexible한 형태를 가지는게 아닐까 추측한다
For the image domain, it is reasonable to consider the spatial hierarchies of lower-level entities and routing can recognize objects similarly to the manner in which we recognize objects.
However, in the language domain,
there is a great deal of freedom in the way that documents and emotions can be expressed.
For example, in the original capsule network, learning to correctly represent the positional characteristics of the eyes, nose, and mouth when categorizing
faces in images was a major challenge.
However, in the case of documents, it is difficult to say that two documents are absolutely different because the order of the sentences in the two documents are different.
In this perspective, it becomes natural to suggest a static routing scheme as follows:

\begin{equation}
\label{eq:staticrouting}
  \begin{aligned}
    s_j = \sum_i W_{ij} h_i \\
    v_j = \frac{||s_j||^2}{1+||s_j||^2}\frac{s_j}{||s_j||},
  \end{aligned}
\end{equation}

where \(W_{ij} = [M \times N]\) is a weight matrix and \(n\) is the number of capsules in \(h_i\).
\(W_{ij}\) is multiplied by \(h_i\) to express the upper entity as a capsule of \(M\)-dimensional vectors. \(v_j\) is the result of applying the squashing function to \(s_j\) and represents the text capsule layer.
This differs from fully connected scalar operations and has the advantage of representing documents as vectors.

\section{Experimental Settings}

\subsection{Datasets}
\begin{table}[ht]
\begin{center}
\resizebox{\columnwidth}{!}{%
   \begin{tabular}{l|r|r|r|r|r|r|r}
   \hline
   Dataset & Classes & Train & Val. & Test & \(|V|\) & \(|V_{pre}|\) & \(l_{avg}\) \\ \hline
   20news & 20 & 10182 & 1132 & 7532 & 177925 & 50021 & 315 \\
   Reuters10 & 10 & 6472 & 720 & 2787 & 28482 & 17508 & 168 \\
   MR (2004) & 2 & 1620 & 180 & 200 & 40693 & 31764 & 779 \\
   MR (2005) & 2 & 8635 & 960 & 1067 & 18764 & 16448 & 22 \\
   TREC-QA & 6 & 4843 & 539 & 500 & 8689 & 7461 & 9 \\
   MPQA & 2 & 8587 & 955 & 1067 & 6246 & 6083 & 3 \\
   IMDb & 2 & 22500 & 2500 & 25000 & 112540 & 58843 & 231 \\
   \hline
   \end{tabular}%
   }
   \end{center}
\caption{\label{tab:datasets} Summary statistics for datasets after tokenization.
%Train, Val., Test: Number of examples per each data partition.
\(|V|\): Vocabulary size. \(|V_{pre}|\): Number of words present in the set of pre-trained word vectors. \(l_{avg}\): Average sequence length.}
\end{table}

We tested our model on seven different benchmark datasets, as shown in \autoref{tab:datasets}.
%Datasets were selected so that document lengths vary.
The details for each dataset are as follows:

\begin{description}
%\footnote{http://qwone.com/~jason/20Newsgroups/}
\item[20news]\footnote{\url{http://scikit-learn.org/0.19/datasets/twenty_newsgroups.html}} This dataset is a collection of 20,000 news documents partitioned between 20 different newsgroups.
%We use the pre-partitioned version provided by the Python machine learning toolkit \textit{scikit-learn}.
\item[Reuters10]\footnote{\url{http://www.nltk.org/book/ch02.html}} We utilize the Reuters corpus provided by the Python natural language toolkit \textit{NLTK}, where documents are initially tagged with 90 categories. In order to limit the number of classes, we selected the 10 most-common categories (earn, acq, money-fix, grain, crude, trad, interest, wheat, ship, corn) and selected corresponding documents.
% If a document is tagged with multiple categories, we designate the first category as its class.
% NOTE (lucypark): 꼼수로 mr2005와 같은 url을 참조하도록 했습니다.
\item[MR (2004)]~\cite{pang2004sentimental}$^3$ A corpus containing 1,000 positive and 1,000 negative preprocessed movie reviews.
\item[MR (2005)]~\cite{pang2005seeing}\footnote{\url{http://www.cs.cornell.edu/people/pabo/movie-review-data/}} A larger movie review dataset, which contains 5,331 positive sentences and 5,331 negative sentences.
%\footnote{\url{http://trec.nist.gov/data/qa.html}}
\item[TREC-QA]~\cite{li2002learning}\footnote{\url{http://cogcomp.org/Data/QA/QC/}} A TREC question dataset for classifying questions into six different question types (person, location, numeric information, etc.).
\item[MPQA]~\cite{wiebe2005annotating}\footnote{\url{http://mpqa.cs.pitt.edu}} Opinion polarity detection of subtasks in the MPQA dataset.
\item[IMDb]~\cite{maas2011learning}\footnote{\url{http://ai.stanford.edu/~amaas/data/sentiment/}} Reviews from the Internet Movie Database, labeled based on positive or negative sentiments.
\end{description}

\subsection{Hyperparameters and training}

\begin{table}[ht]
\begin{center}
\resizebox{\columnwidth}{!}{%
   \begin{tabular}{l|c|c|c|c|c|c|c}
   \hline
   Dataset & $b$ & $l_2$ & $f_n$ & $f_s$ & $lr$ & $a$ & $c_l$ \\ \hline
   20news & 40 & 0.001 & 256 & 5 & 0.001 & 6 & 10/16\\
   Reuters10 & 40 & 0.001 & 256 & 3 & 0.0001  & 6 & 10/16\\
   MR (2004) & 50 & 0.001 & 256 & 3 & 0.001  & 6 & 16/16\\
   MR (2005) & 50 & 0.02 & 256 & 1 & 0.0001  & 16 & 16/24\\
   TREC-QA & 50 & 0.0085 & 256 & 5  & 0.001  & 16 & 32/16\\
   MPQA & 40 & 0.01 & 256 & 1 & 0.00008  & 16 & 8/16\\
   IMDb & 50 & 0.01 & 256 & 6 & 0.001  & 6 & 8/16\\
   \hline
   \end{tabular}%
   }
   \end{center}
\caption{\label{tab:hyperparameter_cap} Hyperparameters used fot the capsule network experiments. $b$: Batch size. $l_2$: Regularization constant in layer $g$ (other layers have a regularization constant of 0.01). $f_n$: Number of filters. $f_s$: Filter size. $lr$: Initial learning rate. $a$: Number of capsules. $c_l$: Dimension of capsules.}
\end{table}

% NOTE(lucypark): explain Y_k... 그런데 왜 margin loss를 사용했냐고 물으면 뭐라고 답할 수 있을까요? Margin loss는 multi-class classification를 위한 것인데 우리의 경우 모든 데이터셋이 단일 클래스이므로 그냥 hinge loss를 사용했다고 하는게 자연스러울 것 같습니다.
% NOTE(lucypark): loss function 얘기는 반박의 여지가 있어서 이번 submission에서는 제외하면 좋을 것 같습니다
% We used \textit{margin loss} as stated in \newcite{sabour2017dynamic},

% \begin{equation}
%   \label{eq:loss}
%   \begin{aligned}
%     L_k &= Y_kmax(0, P)^2 + \frac{1}{2}(1-Y_k) max(0, N)^2\\
%     P &= 0.9-||T_k||, N = ||T_k|| - 0.1.
%   \end{aligned}
% \end{equation}

For training, we utilized preprocessed word vectors consisting of 840 billion words from Glove\footnote{\url{https://nlp.stanford.edu/projects/glove/}}.
We utilized the Adam optimizer~\cite{kingma2014adam}
with exponentially decaying learning rates.
We monotonically decreased the learning rate by decaying it by a factor of 0.99 in every epoch.
We utilized a dropout rate \(p\) of 0.5 and embedding size \(e\) of 300.
% Our model uses pretrained word vectors, consisting of 100 billion words from Google News.\footnote{\url{https://code.google.com/archive/p/word2vec/}}.
% We have experimentally found that using a fixed filter size is more efficient than connecting multiple filter sizes, such as \(f\in(2,3,4,5)\) in our model.
% We used a filter size of \(f\) in datasets.

Particularly, the number of capsules is set to 6, according to experiments based on a held out dataset.
This is a very low number compared to \newcite{sabour2017dynamic}, 
which employed 1,152 capsules for image classification.
Our conjecture for this big difference is that the complexity of the generated feature map is lower in our benchmark tasks.
% FIXME: 아래 문장 clarify
If the complexity of a generated feature map is low, the capsule is expected to provide an appropriate representation of the entity, even without dynamic routing.
% The performance measure we use is \textit{accuracy}

Our model was trained on a GPU utilizing TensorFlow~\cite{abadi2016tensorflow}, with the hyperparameter settings as shown in
\autoref{tab:hyperparameter_cap}.

The CNN classification model from \newcite{kim2014convolutional} was utilized as a baseline model for experimental comparisons.
%We changed the parameters to the same structure.
We performed appropriate parameter tuning for each dataset which are listed in \autoref{tab:hyperparameter_cnn}.

\begin{table}[ht]
\begin{center}
\scalebox{0.73}{%
   \begin{tabularx}{.65\textwidth}{l|X|X|X|X|X}
   \hline
   Dataset & $b$ & $l_2$ & $f_n$ & $f_s$ & $lr$\\
   \hline
   20news & 64 & 0.01 & 256 & [4,5,6] & 0.001 \\
   Reuters10 & 40 & 0.001 & 100 & [2,3,4] & 0.0001  \\
   MR (2004) & 50 & 0.0001 & 256 & [4,5,6] & 0.001  \\
   MR (2005) & 64 & 0.01 & 100 & [2,3,4] & 0.0001  \\
   TREC-QA & 50 & 0.01 & 256 & [3,4,5]  & 0.0001  \\
   MPQA & 64 & 0.01 & 100 & [2,3,4] & 0.0001  \\
   IMDb & 40 & 0.001 & 256 & [3,4,5] & 0.0001  \\
   \hline
   \end{tabularx}%
   }
   \end{center}
\caption{\label{tab:hyperparameter_cnn} Hyperparamers used for the baseline CNN experiments.}
\end{table}

\section{Results and analysis}

\subsection{Classification accuracies}

\begin{table*}[t]\footnotesize
    \renewcommand\arraystretch{1.5}
    \centering
    \begin{tabularx}{\linewidth}{l | l*{7}{X>{\arraybackslash}X}}
        \hline
        Model & 20news & Reuters10 & MR (2004) & MR (2005) & TREC-QA & MPQA & IMDb\\
        \hline
        CapsNet-dynamic-routing & 86.45 & 86.72 & 88.1 & 81.00 & 93.80 & 89.60 & 89.80\\
        CapsNet-static-routing & \textbf{87.17} & \textbf{87.52} & \textbf{89.6} & 80.98 & \textbf{94.84} & \textbf{90.57} & 89.72\\
        \hline
        CNN-non-static* & 86.6 & 87.4 & 88.0 & 81.3 & 92.7 & 89.9 & 90.36\\
        \hline
        CNN-non-static~\cite{kim2014convolutional} & - & - & - & 81.4 & 92.7 & 89.4 & -\\
        DCNN~\cite{kalchbrenner2014convolutional} & - & - & - & - & 93.0 & - &\\
        SA-LSTM~\cite{dai2015semi} & 84.4 & - & - & 80.7 & - & - & 92.76\\
        Virtual adversarial LSTM~\cite{miyato2016adversarial} & - & - & - & \textbf{83.4} & - & - & \textbf{94.1} \\
        Bi-BloSAN~\cite{shen2018bi} & - & - & - & - & \textbf{94.8} & 90.4 & -\\
        \hline
\end{tabularx}
\caption{\label{tab:performance_acc}Text classification accuracies for seven benchmark datasets. Results for CapsNet-* and CNN-non-static are the average accuracies of five consecutive runs. CNN-non-static, marked with an asterisk, are results from a replication code of \protect\newcite{kim2014convolutional}. Other results are from the corresponding references.}
\end{table*}

% As to be seen in the experiments, we found that the performance difference between the two methods was small for smaller datasets, but the larger the dataset size, the better the efficiency of non-routing than the Dynamic routing.
Our experimental results indicate that the accuracy of the static-routing model is higher than that of the dynamic-routing model, as shown in \autoref{tab:performance_acc}.
We believe this is due to the higher complexity of the second layer, which is a feature map utilizing convolutions.
\subsection{Capsule networks over CNNs}

\begin{figure}[ht]
\begin{center}
\includegraphics[scale=0.16]{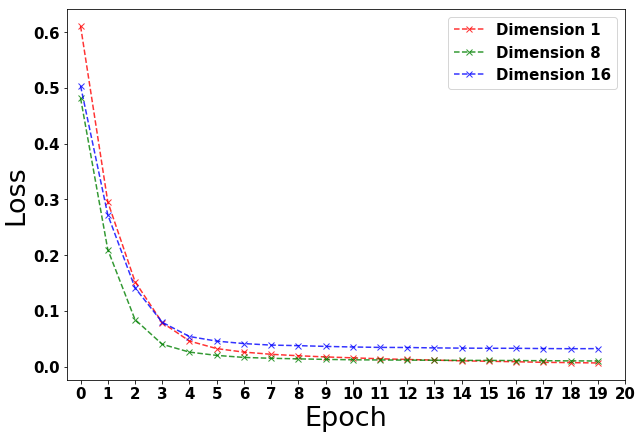}
\includegraphics[scale=0.16]{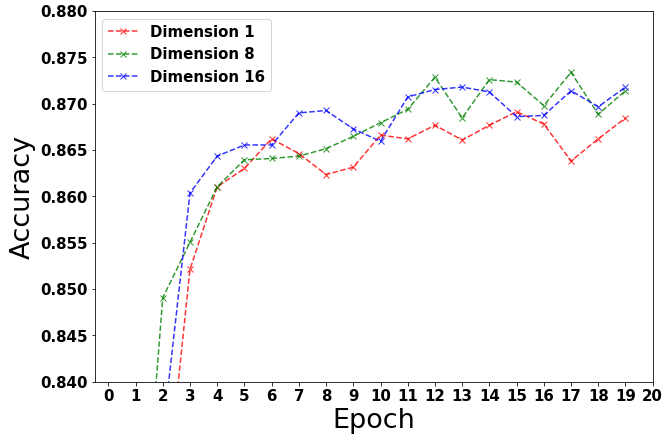}
\caption{Loss and test accuracy with dimensional change of $v_j$ using 20news dataset}\label{fig:train_graph}
\end{center}
\end{figure}
% We use static routing in the text domain.
Static routing does not use all the theoretical philosophies of the capsule network.
However, learning in vector units is different from existing CNN\@.
We experimented with how vector-based learning affects the performance of the model.
\autoref{fig:train_graph} is the performance results according to the variation of $v_j$ dimension but keeping the number of trainable parameters.
Experimental results show that the higher accuracy when the dimension is increased.
Therefore, when training as a vector, the capacity to represent the information of the entities increases and it becomes possible to express various attributes of the entities.
Using static routing does not lose the characteristics of the capsule.
So we experimented with the ability to represent the properties of a capsule in static routing.
We use MNIST because there are some limitations to the visualization of minute changes in words.
We did a perturbation test after adding an ELU-gate in original capsule network structure and changing dynamic routing to static routing.
The experimental method is the same as~\cite{sabour2017dynamic}.

\begin{figure}[ht]
\begin{center}
\includegraphics[scale=0.25]{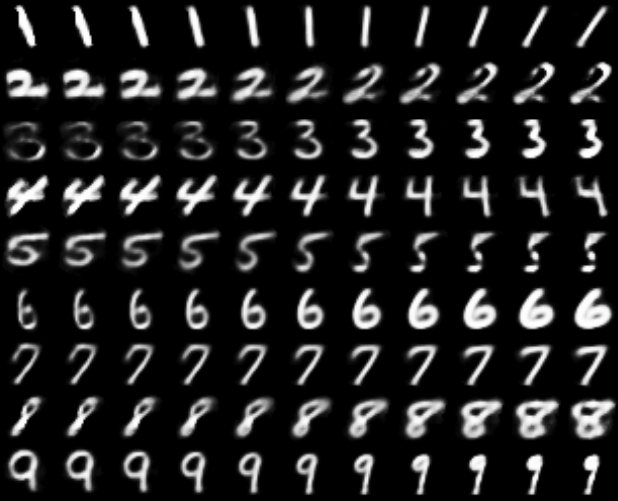}
\caption{Dimension perturbations for static routing model with MNIST. 
%FIXME : 실험 결과의 이유에 대해 추가 설명할 것, 원래 sabour 논문에 있는 MNIST 그림의 의미와 유사하다? sabour의 연구와 동일하게 static routing model을 할 경우에도 MNIST 데이터 셋의 표현에 문제가 없다?
}\label{fig:train_graph}
\end{center}
\end{figure}

\autoref{fig:train_graph} shows that each row has various properties such as rotation, thickness, scale, etc. Therefore, the use of static routing does not lose the essential characteristics of the capsule. This differs from CNN, which is the computation of independent neurons.
% \autoref{fig:train_praph} static routing의 방식도 동일하게 entities 의 properties를 표현할 수 있다는 점은 동일하다. Figure 3 shows that each digit has various attributes. Therefore, the use of static routing does not mean that the essential features of the capsule will disappear.

\begin{table}[ht]
\begin{center}
\scalebox{0.8}{%
\begin{tabular}{c|l|l}
\hline \bf Model & \bf ``good'' & \bf ``bad'' \\ \hline
\hline
\multirow{5}{*}{%
\makecell{Capsule network \\with dynamic routing} } & recommended & waste \\
 & entertaining & worst \\
 & pieces & lame \\
 & gripping & poorly \\
 & truly & disappointing \\
\hline
\multirow{5}{*}{\makecell{Capsule network \\ with static routing} } & delightful & terrible \\
 & refreshing & worst \\
 & pleased & supposed \\
 & fantastic & waste \\
 & terrifying & lame \\
\hline
\multirow{5}{*}{CNN-baseline} & capta & sinny \\
 & 029 & shiksa \\
 & popped & blockbuter \\
 & americanime & u \\
 & waqt & animal \\
\hline
\hline
\end{tabular}
}
\end{center}
\caption{\label{tab:imdb_similarity} Top five neighboring words for ``good'' and ``bad'' in the IMDb dataset, where word vectors were randomly initialized.}
%CapsNet-non-routing(r) Accuracy: 88.21, CapsNet-routing(r) Accuracy: 88.23, CNN(r) Accuracy: 87.60

\end{table}

We measured word similarities to see how our model differs from the basic CNN\@.
% We performed comparisons with a random word vector and learned word vector for additional analysis.
% \autoref{tab:20news_similarity},
\autoref{tab:imdb_similarity} is the similarity measurement table. When the pre-trained word vector was utilized, both the CNN and our model were fine-tuned to the dataset. However, a difference can be seen when utilizing the static-routing method.
In CNN, max-pooling cannot update entire words because only the context with the highest activation is updated during backpropagation.
Because our model does not utilize max pooling, it learns the syntactic representations of words in the static-routing model because its learns without losing positional context.

\subsection{Static-routing over dynamic-routing}\label{sec:static_vs_dynamic}

\begin{table}[ht]
    \begin{center}
    \scalebox{0.73}{%
       \begin{tabularx}{.65\textwidth}{c|X|X|X}
           \hline
           Sentence & Dynamic-routing & Static-routing & Actual class \\ \hline \hline
           \makecell{\textcolor{red}{$<pad>$ $<pad>$ what is the name}\\ \textcolor{red}{of the tallest man of korea}} & 4 & 3 & 3 \\\hline
           \makecell{\textcolor{red}{$Can \, you\, tell\, me$} who is \\ the \textcolor{red}{smallest woman} in \textcolor{red}{usa}} & 2 & 3 & 3 \\\hline
           \makecell{What is \textcolor{red}{the nickname of shakespeare}} & 2 & 3 & 3 \\\hline
           \makecell{\textcolor{red}{what is the name of} the voices of \\the simpsons} & 2 & 3 & 3 \\ \hline
           \makecell{\textcolor{red}{what is the nickname of} \\\textcolor{red}{soccer team of usa}} & 4 & 3 & 3 \\ \hline
           \makecell{\textcolor{red}{can i give a question} , who is the \\ prime minister of \textcolor{red}{norway in europe}} & 4 & 3 & 3 \\ \hline
           \makecell{\textcolor{red}{can you tell me} which president \\ was unmarried \textcolor{red}{in unite states america}} & 2 & 3 & 3 \\ \hline
           \makecell{what is shakespeare \'s nickname} & 3 & 3 & 3 \\\hline
           \makecell{what \textcolor{red}{is the} color \textcolor{red}{of} crickets } & 1 & 2 & 2\\\hline
           \makecell{what are the types of twins} & 4 & 2 & 2\\\hline
           \makecell{\textcolor{red}{i want to ask you} what \\\textcolor{red}{is the public currency of} brazil} & 1 & 1 & 2\\\hline
           %\makecell{\textcolor{red}{do you know the state bird of maryland}} & 1 & 2 \\\hline
           \makecell{\textcolor{red}{can you tell me} what are the \\cigarettes composed of} & 1 & 2 & 2\\ \hline
           \makecell{\textcolor{red}{how we can call female walrus}} & 1 & 2 & 2\\ \hline
           \hline
           \textbf{Pretrained word accuracy} & $64\%$ & $87\%$ & -\\ \hline
           \textbf{Randomly initialized word accuracy} & $65\%$ & $74\%$ & -\\ \hline
           \hline
       \end{tabularx}%
    }
    \end{center}
% \caption{\label{tab:order_test} Sentence used TREC-QA test datas and red color text was changed order, words. Upper tabel ground truth is 3(\texttt{HUMAN}) other is 2(\texttt{ENTITY}). But the sentence means the same label.}
%CapsNet-non-routing(r) Accuracy: 88.21, CapsNet-routing(r) Accuracy: 88.23, CNN(r) Accuracy: 87.60
    % FIXME: "Changed" 라는 모호한 표현을 명확한 표현으로 바꾸기
    \caption{\label{tab:order_test} Sentences from TREC-QA test data where phrases in red have changed word orders. }
\end{table}

% \begin{figure}[ht]
% \begin{center}
% \includegraphics[scale=0.26]{pretrain_correct_c_ij.png}
% \includegraphics[scale=0.26]{pretrain_word_c_ij.png}
% \caption{The degree of coupling of $c_{ij}$ with the upper capsule $v_j$ in the TREC-QA dataset. Above: When test data is not changed. Below: When data is changed.
% The red bar is $v_3$ and ground truth is 3}\label{fig:pretrain_c}
% % \caption{The degree of coupling of $c_{ij}$ toward the upper capsule $v_j$ in the TREC-QA dataset. Above: When test data is not changed. Below: When data is changed.
% % red bar is $v_3$, ground truth is 3} \label{fig:pretrain_c}
% \end{center}
% \end{figure}

It is a general practice to utilize max-pooling in order to extract data features when using a CNN\@.
However, max-pooling often produces poor results in text classification due to loss of information.
More specifically, max pooling only maintains the feature with the highest activation, which means it discards all other features even though they may seemingly be useful.
% in that only fragmented information is categorized because words or phrases excluding the most-activated feature detector are discarded.

% FIXME: hmm... 그런가?
To remedy this issue, capsule networks with dynamic routing chooses to preserve not only one, but all features that are useful, as long as they are ``agreed'' among layers.
However, we assert that this strategy is not necessarily optimal for document classification as opposed to image classification, due to the high variability in text.
% Unlike images, it is somewhat unclear how to define a reasonable positional relationship for lower entities in text.
% Additionally, in dynamic routing, $c_{ij}$, which is based on the cosine similarity between the upper and predicted capsules, determines the degree of coupling between the upper capsules. 
Specifically, the model should be flexible and robust enough to handle slight modifications in the text,
such as word order shuffling or the insertion of an untrained word vector.
We conjecture that removing the coupling coefficient would smooth out the underlying signals and therefore make the model more robust in this regard. We further perform experiments involving word order shuffling and noise injections in Section \ref{sec:static_vs_dynamic}
to support this claim.
% Therefore, for text, we would not be confident in the coupling of the upper capsules.

% To remedy this issue, dynamic routing is utilized for capsule networks. In dynamic routing, $c_{ij}$, which is based on the cosine similarity between the upper and predicted capsules, determines the degree of coupling between the upper capsules. So capsules with different directions from upper capsules are down scaled and expressed as the sum of vectors, so the upper capsule has a strong confident. However, since the text has a high degree of freedom of expression, the same document can be expressed in various ways, so the ability to generalize the model is important. Therefore, rather than scaling the amount of information to be sent to the upper capsule, using all of the capsule information may be more effective in terms of generalization of the model. Since the existence of entities is determined through the squashing before the routing, the expression of the upper capsule is not changed because of the low existence of the capsule, and the norm of the upper capsule can be smoothed by adding the vector of all the capsules.

% FIXME: 위에서는 (3.1.2) sentence의 배열이 바뀌어도 된다고 하고 실험은 단어 배열이 바뀌는 것으로 하면 안될 것 같습니다.
In order to prove the above hypothesis,
and argue the effectiveness of static routing,
we evaluated the classification results after changing the sequences of words in a sentence.
For this, we utilized 50 samples from each of class 2 (\texttt{ENTITY}) and 3 (\texttt{HUMAN}), from the TREC-QA test dataset.
As can be seen in \autoref{tab:order_test}, static routing proved a much higher accuracy compared to dynamic routing, given that the word vectors are pretrained.
% FIXME: static routing의 결과가 맞나요? dynamic routing과 비교해야하는건 아닌가요? 이 그래프가 굳이 필요할까요? 필요하다면 더 상세한 설명이 들어가야할 것 같습니다. 일단 지우겠습니다.
% One can also see from \autoref{fig:pretrain_c} that the x-axis is the number of capsules and the y-axis is the weights of $c_{ij}$, which determines the coupling of the upper capsules.
% % The results vary based on changes in the document.
% This is because the directionality of the vectors of capsules changes. In this case, the coupling coefficient does not impart a weight to the correct upper capsules.

\begin{figure}[htb!]
\centering
\begin{subfigure}[b]{0.35\textwidth}
  \includegraphics[width=1\linewidth]{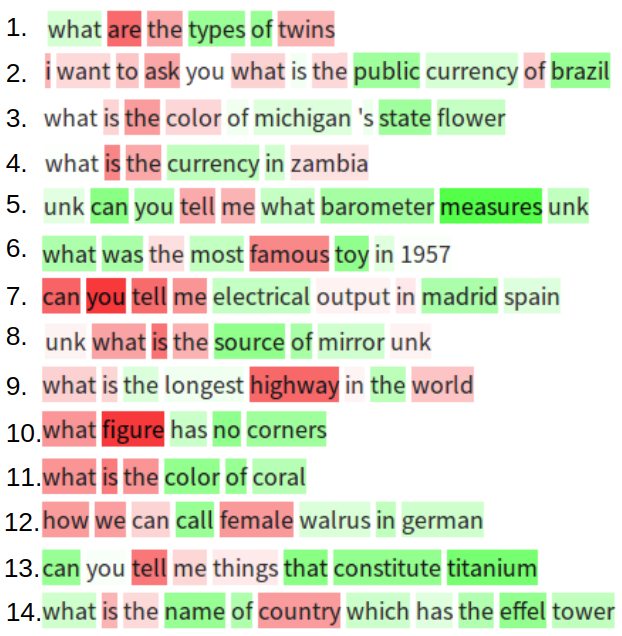}
  \subcaption{Dynamic routing}\label{fig:word_weight_routing}
\end{subfigure}
\begin{subfigure}[b]{0.35\textwidth}
  \includegraphics[width=1\linewidth]{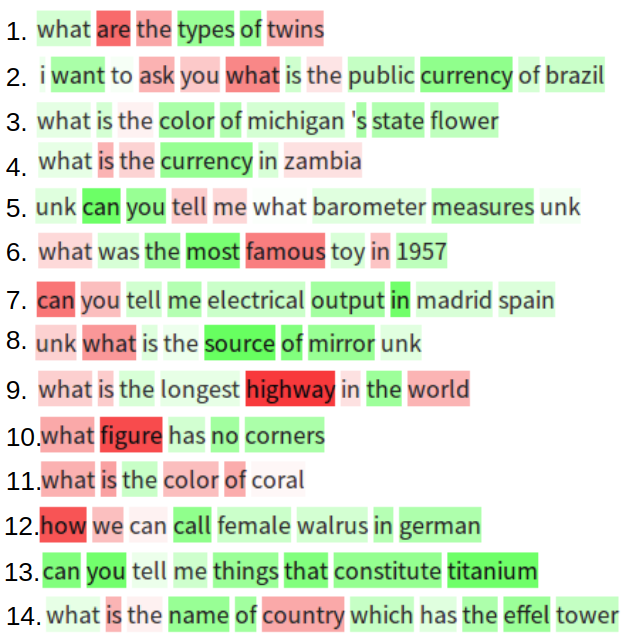}
  \subcaption{Static routing}\label{fig:word_weight_rand_routing}
\end{subfigure}
\caption{Words are highlighted according to their importance for prediction for the TREC-QA dataset, where the ground truth class is 2 (\texttt{ENTITY}). Green and red labels are positive and negative effects, respectively. Higher intensity indicates greater importance of the word.}\label{fig:degree_coupling}
\end{figure}

% 단어가 미치는 영향
% FIXME: lime의 결과가 무엇을 의미하는 것일까요?
We further identified the effects of words changes on the predictions of the model utilizing LIME~\cite{ribeiroshould}.
LIME is a method for generating new samples with similar values to corresponding instances in the vicinity of the predicted value from the model and determining how the predictions of the models differ based on the input values.
% In addition to the incorrect decision of the coefficient, we have identified the effect of the word on the prediction of the model using LIME\cite{ribeiroshould}
% LIME is a method for generating new samples with similar values to the corresponding instances in the vicinity of the prediction value of the model and determining how the predictions of the models differ according to the input values.
% 우리는 LIME 알고리즘을 사용했다. LIME(Local Interpretable Model-agnostic Explainations)은 모델의 Prediction 값 근방에서 해당 인스턴스와 유사하지만 조금 값이 다른 새로운 샘플들을 만들어 내고 이 샘플들에 대해 모형이 내놓는 예측값이 input 값에 따라 어떻게 다른지 파악하는 방식이다.
% In this case, we compared the methods of dynamic routing and static routing.
In the results presented in \autoref{fig:degree_coupling},
% the green highlights indicate a positive effect on prediction. The red highlights indicate a negative effect on prediction.
% FIXME: 둘다 잘못하지만 dynamic routing이 더 잘못한다는 얘기를 하고 싶은 것인가?
both routing models tend to produce incorrect decisions because of changed words.

% \begin{figure}[ht]
% \begin{center}
% \includegraphics[scale=0.3]{paper_works/EMNLP/non-routing-word.png}
% \caption{rand-routing model, ground truth 2(ENT).} \label{fig:word_weight_rand_routing}

% \includegraphics[scale=0.3]{paper_works/EMNLP/routing-word.png}
% \caption{routing mode, ground truth 2(ENTITY)} \label{fig:word_weight_routing}
% \end{center}
% \end{figure}

% FIXME: reconstruction 얘기가 갑자기 나오는 느낌입니다.
When the original example is ``what is the state flower of michigan'' (the third example in \autoref{fig:word_weight_routing}), the reconstructed data is ``what is the color of michigan's state flower''.
% FIXME: 아래 문장... so what?
The dynamic routing method has a negative effect on the newly added word ``color.''
% FIXME: 역시 so what?
It also has a negative effect when a combination that does not appear in the existing TREC-QA data is added, such as ``can you tell me'' in the table (second to last example in \autoref{fig:word_weight_routing}).

% The result as shown in \autoref{fig:word_weight_routing}, the green highlight has a positive effect on the prediction. The red highlight has a negative effect on the prediction.
% The routing model tend to wrong decision because of changed words.

% When the existing data of TREC-QA is ``what is the state flower of michigan'' and the reconstructed data is ``what is the color of michigan 's state flower''. The dynamic routing method has a negative effect on the newly added ``color''. It also has a negative effect when a combination that does not appear in the existing TREC-QA data is added, such as ``can you tell me'' in the \autoref{fig:word_weight_routing}.

% 기존의 데이터가 what is the state flower of michigan, 재구성한 데이터가 what is the color of michigan 's state flower 이라고 할 때 , dynamic routing 방식이 새롭게 추가된 color에 대해서 negative 영향을 끼쳤다. 또한 표에서 보여지는 can you tell me 처럼 기존의 TREC data에는 등장하지 않는 조합이 추가될 경우에도 negative 영향을 가지고 왔다.
% Therefore, we did not use the coupling factor for this reason. As a result, the complexity of the computational complexity can be reduced and the generalization is better than when routing is used.
% FIXME: static routing is used? dynamic routing is used 가 아니라? => 수정했습니다.
Therefore, we did not utilize the coupling factor for this reason.
As a result, the computational complexity can be reduced and generalization is improved compared to when dynamic routing is used.

\subsection{Justifying the ELU-gate}

\begin{table}[ht]
\begin{center}
\scalebox{0.73}{%
   \begin{tabularx}{.65\textwidth}{l|X|X|X|X}
   \hline
   Dataset & TREC-QA & MPQA & 20news &  MR (2004)\\ \hline\hline
   \makecell{Multiple filters \\ without max-pooling} & 93.47 & 88.68 & 85.76 & 85.89  \\\hline
   \makecell{Multiple filters \\ with max-pooling} & 93.84 & 89.28 & 86.08 & 85.69  \\ \hline
   Convolutional layer & 93.99 & 90.07 & 85.40 & 85.29   \\ \hline
   ELU-gate & 94.80 & 90.57 & 87.17 & 89.60  \\\hline
   \hline
   \end{tabularx}%
   }
   \subcaption{Static-routing}
   \end{center}
\begin{center}
\scalebox{0.73}{%
   \begin{tabularx}{.65\textwidth}{l|X|X|X|X}
   \hline
   Dataset & TREC-QA & MPQA & 20news &  MR (2004)\\ \hline\hline
   \makecell{Multiple filters \\ without max-pooling} & 91.95  & 88.82  & 85.56  & 84.59  \\\hline
   \makecell{Multiple filters \\ with max-pooling} & 92.63  & 89.43  & 85.69 & 87.29  \\ \hline
   Convolutional layer & 93.07  & 90.26 & 85.60 & 85.29   \\ \hline
   ELU-gate & 93.80  & 89.60  & 86.45  & 88.10  \\\hline
   \hline
   \end{tabularx}%
   }
   \subcaption{Dynamic-routing}
   \end{center}
\caption{\label{tab:ablation_test} Results of the ablation test. Each number is the mean accuracy of five consecutive runs.}
\end{table}

% In this section, we perform an ablation study of the impact of ELU-gate.
In~\newcite{dauphin2016language}, the ELU gating mechanism was mainly experimented with recurrent models such as LSTM and GRU,
but they also showed that it is effective with convolutional layers.
% FIXME: 아래 문장들이 무슨 뜻이죠? 일단 흐름상 어색해서 지웠습니다.
% It is important to fine-tune the pretrained word vector in classification.
% Therefore, when the effect of the gradient is downscaled during back propagation, proper fine-tuning is not performed.
The gate gradient of LSTM is as follows.

\begin{equation}
    \begin{aligned}
        & \nabla[ \tanh(X)  \otimes \sigma(X)] =\\
        & \tanh'(X)\nabla X \otimes \sigma(X) + \tanh(X) \otimes \sigma'(X)\nabla X
    \end{aligned}
\end{equation}

In the case of LSTM, the effect of the gradient is reduced because downscaling occurs in
$\tanh'(X)$ and $\sigma'(X)$.

\begin{equation}
\label{eq:elu_gradient}
\begin{aligned}
    & \nabla[ X  \otimes elu(X)] =\\
    & \nabla X \otimes elu(X) + X \otimes elu'(X) \nabla(X)
\end{aligned}
\end{equation}

Since the gradient of the ELU-gate can be expressed as shown in \autoref{eq:elu_gradient}, the effect of downscaling is small.
Unlike max-pooling, fine-tuning works well because input words are updated globally.
%설명 추가하기

\autoref{tab:ablation_test} shows the results of comparing the accuracy with ELU-gate and other structures.
The multiple filter layer is a convolution layer having a filter size of $[3, 4, 5]$ as in the case of CNN~\cite{kim2014convolutional} structure.
The number of filters in the multiple filter layers was 100 per filter, and the kernel size of pooling was $[2\times1]$.
% FIXME: clarify
The convolutional layer is a layer that excepted ELU-gate.

\subsection{Text transformation}
% FIXME: 아래 실험의 motivation이 뭔지 얘기해줘야할 것 같습니다. 결과만 있고 해석이 없기 떄문에 regularization effect를 기대한것인지, 유의미한 manifold를 찾았다고 주장하고 싶은것인지, 등 여러 가지 추측을 하게 됩니다.

% 기존의 캡슐 네트워크가 여러 properties를 바탕으로 아핀변환된 entities를 표현하는 실험을 함으로써 캡슐이 properties를 나타낼 수 있다는 것에 설득력을 보태주었듯이 저희도 문서에서 변환된 형태를 보여주어야 좀 더 설득력이 생길 것 같았어요.
% 이미지에서도 entities의 변환은 주성분(entities에는 없어서는 안되는 주요 특성, 예를 들어 6을 표현하기 위해서는 이것만큼은 꼭 있어야 하는 특성)은 유지한채 다양한 변환을 한다고 생각했고 그렇다면 문서에서도 동일하게 적용될 것이라고 생각하여 실험을 했습니다! 따라서 각 문서에서 가장 뼈대가 되는 단어들(표를 예를 들자면 what, name)의 변화보다는 좀 더 세부적인 특징(예: 이름)들에 대해서 다양성을 줄 것이라는 가정을 했었습니다.
% 따라서 문서의 reasonable transformation은 메인이 되는 의미는 유지하면서 다양한 특징들을 변환할 수 있겠다고 생각해서 실험을 하게 되었습니다. (이 부분을 영어로 쓰면 저의 형편없는 영어실력때문에 전달력이 떨어질 것 같아서 한글로 작성했어요...죄송합니다ㅠ)

In image classification, capsules represent the various properties of a particular entity that is present in an image.
These properties include types such as tilt, orient, hue, etc.
% Thus, capsules in an image can represent affine transformed entities.
In order to apply this analogy to text,
% FIXME: converted?
% Text의 단어가 바뀌었다는 뜻입니다....ㅠ
we experimented with documents to see how capsules can learn the innate characteristics of the document being converted.
To test this reconstructive phenomena, we added three fully connected layers to the capsule network with static-routing.

\begin{figure}[ht]
\begin{center}
\includegraphics[scale=0.26]{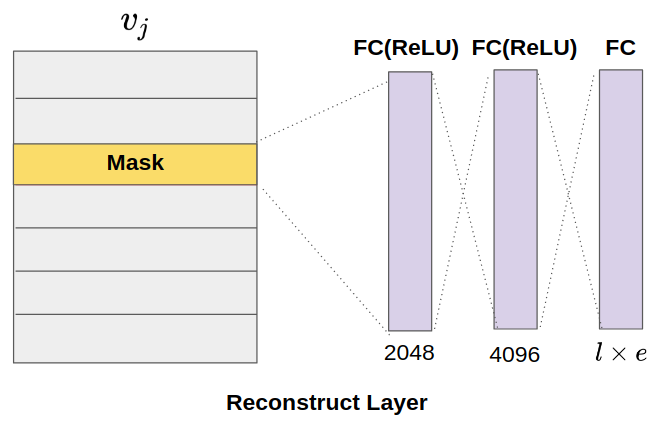}
\caption{\label{fig:reconstruct} Reconstruct layer consisting of 3 fully connected layers, $l$ is sequence length, $e$ is embedding size.}
\end{center}
\end{figure}

We added the MSE loss between the input and the reconstruct layer output, and downscaled the MSE loss by 0.03.
Pretrained word vectors were not utilized.
We confirmed the decoder results when we gave random noise between -0.3 and 0.3 to each dimension of the activated capsule in $v_j$.
We used the words with the highest value by measuring the cosine similarity of each row and vocabulary.

\begin{table}[ht]
\begin{center}
\scalebox{0.73}{%
   \begin{tabularx}{.65\textwidth}{c|X|X}
   \hline
   Sentence & Noise & i \\ \hline \hline
   \makecell{what is the name of neil armstrong's wife} & 0 & - \\\hline
   \makecell{what is the name of \textcolor{blue}{john davis's one father}} & 0.3 & 1 \\ \hline
   \makecell{what is the name of \textcolor{blue}{john george's one father}} & -0.3 & 2 \\ \hline
   \makecell{what is the name of \textcolor{blue}{richard davis's one}} & 0.3 & 2 \\ \hline
   \makecell{what is the name of \textcolor{blue}{richard davis's one father}} & -0.2 & 4 \\ \hline
   \makecell{what \textcolor{blue}{was} the name of \textcolor{blue}{john davis's one father}} & -0.3 & 5 \\ \hline
   \makecell{what is the name \textcolor{blue}{that richard davis's one sons}} & 0.3 & 7 \\ \hline
   \makecell{what \textcolor{blue}{was} the name \textcolor{blue}{of john davis's one}} & -0.3 & 8 \\ \hline
   \makecell{what is the name \textcolor{blue}{of richard davis's one daughters}} & 0.3 & 10 \\ \hline
   \makecell{what \textcolor{blue}{was} the name \textcolor{blue}{of richard} \\\textcolor{blue}{robinson's one daughters}} & 0.3 & 12 \\ \hline
   \makecell{what is the \textcolor{blue}{that of one american there}} & -0.2 & 12 \\ \hline
   \makecell{what \textcolor{blue}{was} the name of \textcolor{blue}{john davis's one daughters}} & 0.2 & 15 \\ \hline
   \hline
   \end{tabularx}%
   }
   \end{center}
\caption{\label{tab:transformation}
Random noise is added to the $i$-th element of the 16-dimensional vector.}
\end{table}

The first row in \autoref{tab:transformation} is the original sentence of TREC-QA with no added noise.
When the noise is added, the result does not change the meaning of the question, but some words changed.
Also, the changed sentence is a newly created without the same as the dataset.
In the case of words, we could not visualize detailed changes like images because measured the similarity of the words included in the vocabulary.

\section{Conclusion}

In this paper, we proposed the application of capsule networks to the text classification domain and suggested the utilization of a static routing variant.
We compared the proposed model to CNNs, and demonstrated that capsule networks are indeed useful for text classification based on seven popular benchmark datasets.
We additionally proposed static routing, an alternative to dynamic routing, that results in higher classification accuracies with less computation.
% We also observed the behavior of coupling coefficients over several routing iterations and calculated word similarities to demonstrate that the proposed method produces meaningful text representations.

%showed the advantages of the capsule through experiments.

% \section*{Acknowledgments}

% The acknowledgments should go immediately before the references.  Do
% not number the acknowledgments section. Do not include this section
% when submitting your paper for review. \\

%\noindent {\bf Preparing References:} \\

% Include your own bib file like this:
% {\small\verb|\bibliographystyle{acl_natbib_nourl}|
% \verb|\bibliography{emnlp2018}|}

% Where \verb|emnlp2018| corresponds to the {\tt emnlp2018.bib} file.
\bibliography{text_capsnet.bib}
\bibliographystyle{acl_natbib_nourl}

\newpage

\end{document}